\documentclass[letterpaper]{article} 
\usepackage{aaai2026}
\usepackage{times}  
\usepackage{helvet}  
\usepackage{courier}  
\usepackage[hyphens]{url}  
\usepackage{graphicx} 
\usepackage{natbib}  
\usepackage{caption} 
\usepackage{algorithm}
\usepackage{algorithmic}
\usepackage{amsfonts} 
\usepackage{amsmath}
\usepackage{booktabs}
\usepackage{xcolor}
\usepackage{graphicx}
\usepackage{caption}
\usepackage{multicol}
\usepackage{multirow}
\usepackage{newfloat}
\usepackage{listings}
\usepackage{makecell}
\usepackage{xspace}
\newcommand{\approach}{DocR1\xspace}
\newcommand{\AnnotationData}{ArxivFullQA\xspace}
\newcommand{\TrainData}{EviBench\xspace}

\DeclareCaptionStyle{ruled}{labelfont=normalfont,labelsep=colon,strut=off} 
\floatstyle{ruled}
\newfloat{listing}{tb}{lst}{}
\floatname{listing}{Listing}
\nocopyright

\title{DocR1: Evidence Page-Guided GRPO for Multi-Page Document Understanding}

\author{
\textbf{Junyu Xiong}\textsuperscript{\rm 1}\equalcontrib\quad
\textbf{Yonghui Wang}\textsuperscript{\rm 1}\equalcontrib\quad
\textbf{Weichao Zhao}\textsuperscript{\rm 1}\quad
\textbf{Chenyu Liu}\textsuperscript{\rm 2} \\
\textbf{Bing Yin}\textsuperscript{\rm 2}\quad
\textbf{Wengang Zhou}\textsuperscript{\rm 1}\thanks{Corresponding author.}\quad
\textbf{Houqiang Li}\textsuperscript{\rm 1}
}

\affiliations{
\textsuperscript{\rm 1} University of Science and Technology of China  \quad
\textsuperscript{\rm 2} iFLYTEK Research \\
xiongjyu@mail.ustc.edu.cn,\; wyh1998@mail.ustc.edu.cn,\; zhwg@ustc.edu.cn
}

\usepackage{bibentry}

\begin{document}

\maketitle

\begin{abstract}
Understanding multi-page documents poses a significant challenge for multimodal large language models (MLLMs), as it requires fine-grained visual comprehension and multi-hop reasoning across pages. While prior work has explored reinforcement learning (RL) for enhancing advanced reasoning in MLLMs, its application to multi-page document understanding remains underexplored. In this paper, we introduce \textbf{DocR1}, an MLLM trained with a novel RL framework, \textbf{Evidence Page-Guided GRPO (EviGRPO)}. EviGRPO incorporates an evidence-aware reward mechanism that promotes a coarse-to-fine reasoning strategy, guiding the model to first retrieve relevant pages before generating answers.
This training paradigm enables us to build high-quality models with limited supervision.
To support this, we design a two-stage annotation pipeline and a curriculum learning strategy, based on which we construct two datasets: \TrainData, a high-quality training set with 4.8k examples, and \AnnotationData, an evaluation benchmark with 8.6k QA pairs based on scientific papers. Extensive experiments across a wide range of benchmarks demonstrate that DocR1 achieves state-of-the-art performance on multi-page tasks, while consistently maintaining strong results on single-page benchmarks.
\end{abstract}

\section{Introduction}
Documents are ubiquitous in everyday life, encompassing scanned forms, tables, charts, and PDFs. Consequently, document understanding is a critical capability for multimodal large language models (MLLMs). While recent MLLMs have achieved strong performance on single-page document tasks~\cite{hu2024mplug1.5, zhou2024doge, liu2024textmonkey, wang2023towards, zhao2024tabpedia, feng2024docpedia,Xia_2024, dong2025spreadsheetllmencodingspreadsheetslarge}, these tasks often cover a subset of real-world applications. In practice, many tasks involve multi-page documents—such as scientific papers, contracts, and reports—which require not only fine-grained visual understanding but also the ability to retrieve relevant pages and reason across dispersed content.

\begin{figure}[!t]
\centering
\includegraphics[width=\linewidth]{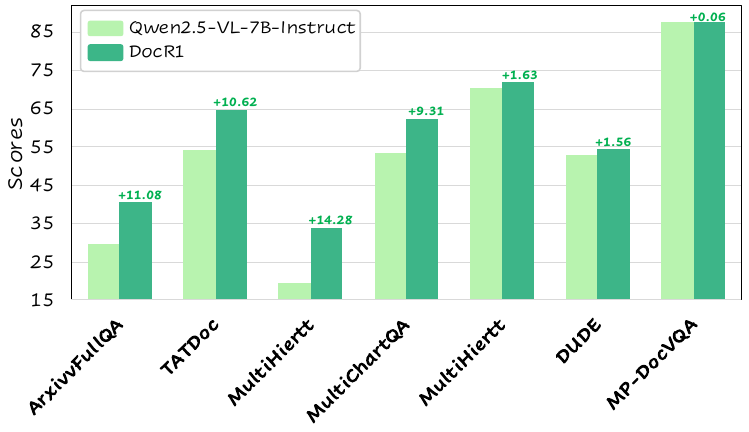}
\caption{Our \approach has significant improvements on various benchmarks of multi-page document understanding.}
\label{fig:fig-intro}
\end{figure}

Recent advances, including OpenAI’s o1~\cite{jaech2024openai} and DeepSeek-R1~\cite{guo2025deepseek}, highlight the growing importance of advanced reasoning in LLMs. Parallel work has begun to explore reinforcement learning (RL) in MLLMs. For example, VisualRFT~\cite{liu2025visual} introduced verifiable reward functions for perception tasks and applied GRPO~\cite{shao2024deepseekmath} to enhance object detection and visual classification. Similarly, Vision-R1~\cite{huang2025vision} integrated GRPO with a PTST training strategy to improve mathematical reasoning capabilities.
However, these methods are restricted to single-image inputs and focus on either perception or symbolic reasoning. For multi-image tasks, methods like Video-R1~\cite{feng2025video} and VideoChat-R1~\cite{li2025videochat} have applied GRPO in the video domain, incorporating customized reward functions to capture spatiotemporal dependencies. Nevertheless, applying RL to multi-page document understanding—where effective evidence retrieval and multi-hop reasoning are both critical—remains largely unexplored.

To bridge this gap, we propose \textbf{DocR1}, a novel method equipped with the \textbf{Evidence Page-Guided GRPO (EviGRPO)} RL framework, specifically designed to enhance multi-page reasoning in MLLMs.
EviGRPO extends the GRPO paradigm by introducing an evidence-aware reward mechanism that promotes a human-like, coarse-to-fine reasoning process: first forming a global understanding of the document, then retrieving the relevant pages, and finally reasoning over them to derive the answer. Specifically, we define three verifiable rewards to guide model optimization: format consistency, evidence page accuracy, and answer accuracy.
To support this process, we design a rigorous two-stage annotation pipeline comprising a generation step and a verification step. 
In the generation step, an MLLM is prompted to produce initial annotations. 
In the verification step, the same MLLM is prompted to validate the generated annotations, ensuring higher data quality through self-checking.
Using this pipeline, we construct two datasets: \TrainData, a minimally supervised yet high-quality training set containing 1.3k single-page and 3.5k multi-page samples; and \AnnotationData, a new benchmark with 8.6k QA samples designed to evaluate document-level reasoning over full scientific papers.
In addition, we introduce a two-stage curriculum training strategy to enable more effective learning under limited supervision.
The model is first trained on single-page data to internalize the desired output format and reasoning style, and then trained on multi-page data to develop its multi-page reasoning capabilities.

Our proposed \approach is an MLLM tailored for complex document-level reasoning.
Rather than producing only final answers, \approach explicitly outputs intermediate reasoning traces and evidence localization, thereby improving both interpretability and reliability.
As shown in Figure~\ref{fig:model-overview}, the model follows a structured decision-making path aligned with human reading behavior.
Comprehensive evaluations across both multi-page and single-page benchmarks validate the effectiveness of our method. 
Specifically, \approach achieves state-of-the-art performance on multi-page benchmarks, attaining an average score of 59.36, outperforming the baseline by 6.93 points.

Our contributions can be summarized as follows:
\begin{itemize}
\item We present \approach, an MLLM designed to generate structured outputs comprising the models's thought process, selected evidence pages, and final answers.
\item We propose EviGRPO, an RL framework specifically designed for multi-page document understanding in MLLMs. EviGRPO adopts a coarse-to-fine reasoning strategy, where the model first identifies the relevant evidence pages before generating an answer.
\item We develop a two-stage annotation pipeline and a curriculum training strategy that together enable high-quality data labeling and efficient model training under limited supervision.
\end{itemize}
\section{Related Work}
\subsection{Document Understanding with MLLMs}
MLLMs have recently advanced document understanding by enabling unified modeling of textual and visual information, without relying on traditional OCR engines. Several recent works have explored MLLM-based document understanding. UReader~\cite{ye2023ureader} pioneered OCR-free, end-to-end instruction tuning across multiple tasks. 
The mPLUG-DocOwl series~\cite{ye2023mplug, hu2024mplug1.5, hu2024mplug2} improved structure awareness and computational efficiency through structural modeling and visual token compression. TextMonkey~\cite{liu2024textmonkey} introduced shifted window attention to preserve semantic continuity across image patches. DOGE~\cite{zhou2024doge} proposed a document-oriented grounding framework with high-quality training data, while Doc-CoB~\cite{mo2025doc} enabled dynamic region selection for focused, step-by-step reasoning.  However, most methods are still limited to single-page documents. Our work targets more complex multi-page document tasks in real scenarios.

\subsection{Reinforcement Learning for MLLMs}

RL has shown promise in enhancing the reasoning capabilities of LLMs. While early approaches relied on pretrained reward models, recent work such as DeepSeek-R1~\cite{guo2025deepseek} demonstrates that simple, rule-based rewards can provide scalable and verifiable supervision. Building on this insight, researchers have extended RL to multimodal large language models (MLLMs). VisualRFT~\cite{liu2025visual} introduced visually grounded, verifiable reward functions for fine-grained classification and few-shot detection using GRPO. Vision-R1~\cite{huang2025vision} further leveraged GRPO with a PTST strategy to improve mathematical reasoning on a curated multimodal dataset. In the video domain, Video-R1~\cite{feng2025video} proposed T-GRPO to explicitly incorporate temporal cues during training, marking the first systematic application of GRPO to video-based MLLMs. VideoChat-R1~\cite{li2025videochat} also applied GRPO to enhance spatiotemporal reasoning, achieving strong performance on temporal grounding and tracking tasks. While these studies focus on perception and video reasoning, RL has not yet been explored for multi-page document understanding. Our work aims to fill this gap.

\section{Method}

\begin{figure*}[!t]
\centering
\includegraphics[width=\textwidth]{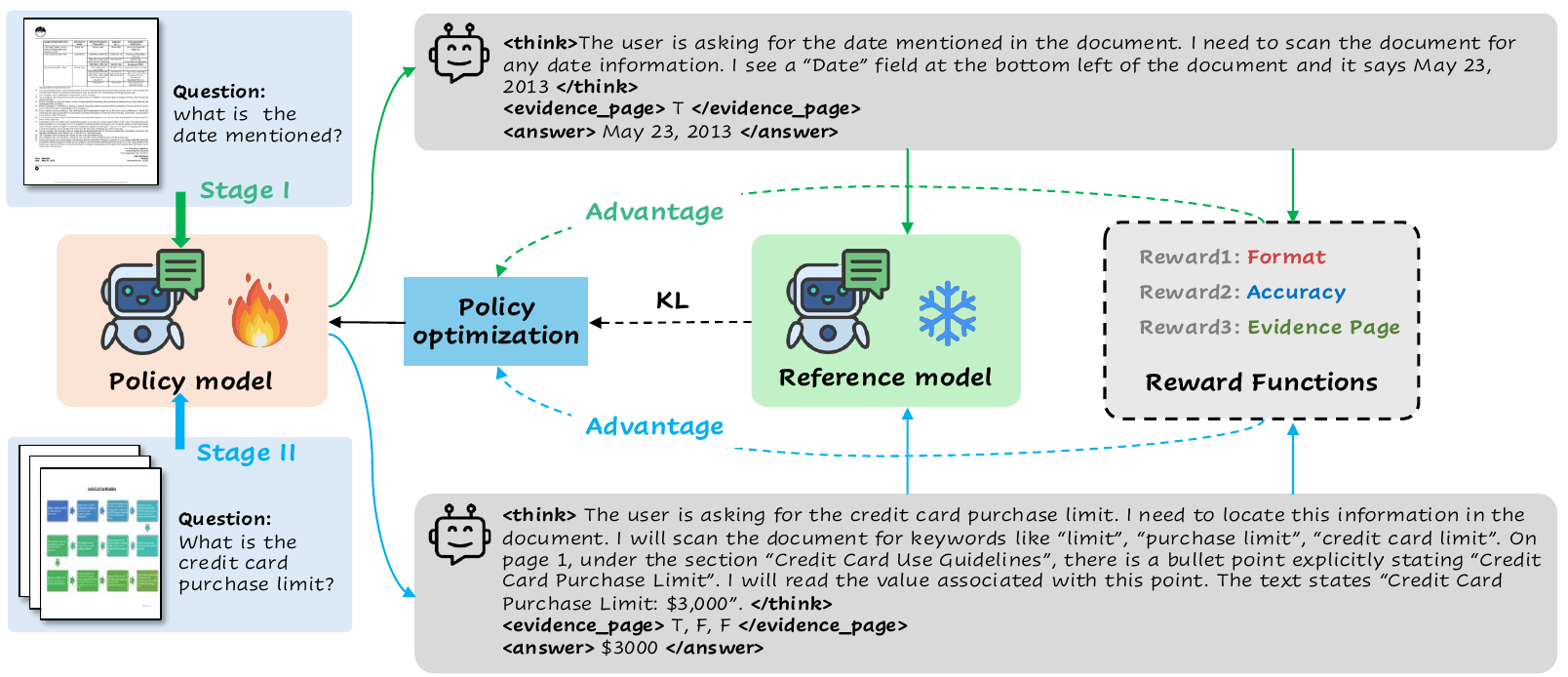}
\caption{Our proposed EviGRPO training framework adopts a two-stage strategy to progressively enhance the model's multi-page reasoning capabilities.}
\label{fig:model-overview}
\end{figure*}

\subsection{EviGRPO}
While the GRPO algorithm~\cite{shao2024deepseekmath} has demonstrated promise in enhancing reasoning capabilities, its direct application to multi-page document understanding remains suboptimal. To address this gap, we introduce \textbf{Evidence Page-Guided GRPO (EviGRPO)}, a variant of GRPO tailored specifically for multi-page document tasks, as illustrated in Figure~\ref{fig:model-overview}.

When engaging in multi-page reading comprehension, humans typically begin by identifying the pages likely to contain the answer, and then focus on locating the specific regions that correspond to the question and answer within those pages.
Inspired by this ``coarse-to-fine'' reading strategy, EviGRPO mimics the human approach by first selecting a small set of potentially relevant pages at a coarse level, followed by fine-grained reasoning over the selected content.
This hierarchical reading paradigm facilitates a more efficient and accurate understanding of multi-page documents.

EviGRPO builds upon the original GRPO structure, which includes format and accuracy rewards, by introducing an additional evidence-aware reward that encourages grounding answers on relevant document pages.
For each training sample, consisting of a question \( q \) and \( N \) document images, EviGRPO first generate \( G \) candidate responses \( \mathcal{O} = \{o_1, o_2, \ldots, o_G\} \) using the current policy.  
For each candidate response \( o_i \), the verifiable total reward \( r_i \) is defined as the sum of three components:
\begin{equation}
r_i = r_i^{\text{format}} + r_i^{\text{acc}} + r_i^{\text{evi}}.
\label{eq:total_reward}
\end{equation}
where \( r_i^{\text{format}} = 1 \) if the model's output strictly adheres to the formatting rules specified in Table~\ref{system prompt 3 ours}, and \( r_i^{\text{format}} = 0 \) otherwise.
The accuracy reward \( r_i^{\text{acc}} \) is defined as the ANLS score between the model-generated answer and the ground-truth.
The evidence-page reward \( r_i^{\text{evi}} \) measures whether the model correctly identifies the supporting pages. 
Let \( N_i' \) denote the number of predicted evidence pages, and let \( P_i \) and \( G_i \) represent the sets of predicted and ground-truth evidence pages, respectively.
We define \( r_i^{\text{evi}} \) as the F1-style overlap between the predicted and ground-truth evidence page sets.
Specifically, it is computed as:
\begin{equation}
r_i^{\text{evi}} = 
\begin{cases}
\displaystyle \frac{2 \cdot |P_i \cap G_i|}{|P_i| + |G_i|}, & \text{if } |P_i| + |G_i| > 0  \ \ and \ \ N = N_i' \\
0, & \text{otherwise}
\end{cases}
.
\label{eq:reward_evi}
\end{equation}

\begin{table}[!t]
\centering
\begin{tabular}{p{\linewidth}}
\toprule
\textbf{System:} 
You will be given one or more images along with a question. Your task is to understand the visual content and answer the question.  First, think carefully about the question and present your reasoning in \texttt{<think>} and \texttt{</think>}.  Next, identify how many pages (images) are provided, and for each page, determine whether it contains relevant evidence to answer the question.  List your judgments in \texttt{<evidence\_page>} and \texttt{</evidence\_page>} using a comma-separated sequence of T (True) or F (False), one for each page, in order (e.g., T, F, T, F).  Finally, provide your answer in \texttt{<answer>} and \texttt{</answer>}. The answer should be one or more words or phrases.

\textbf{User:} {\texttt{prompt}}. \textbf{Assistant:} \\
\bottomrule
\end{tabular}
\caption{The page selection format of EviGRPO. {\texttt{prompt}} will be replaced with the specific question.}
\label{system prompt 3 ours}
\end{table}

We adopt the F1 score rather than accuracy to mitigate reward hacking.
In multi-page documents, where evidence pages are typically sparse, predicting all pages as irrelevant can lead to artifically high accuracy.
The F1-based reward addresses this issue by balancing precision and recall.
To further encourage fine-grained reasoning, the model is required to assess each image individually, labeling each as either relevant (T) or irrelevant (F).
To enforce this behavior, we set the reward to zero if the number of predicted judgments \(N_i'\) does not match the total number of input images \(N\).
Next, following the GRPO framework, EviGRPO computes the mean and standard deviation of the total rewards across the response set \( \mathcal{O}\) , and normalizes each individual reward to obtain its corresponding advantage value \( A_i \):
\begin{equation}
A_i = \frac{r_i - \text{mean}(\{r_i\}_{i=1}^{G})}{\text{std}(\{r_i\}_{i=1}^{G})}.
\label{eq:evi_advantage}
\end{equation}

The final policy optimization objective is to maximize the expected return.
To regularize the update, a KL divergence term \(\mathrm{D}_{\mathrm{KL}}(\cdot \,\|\, \cdot)\) is introduced to constrain the optimized policy \(\pi_\theta\) from diverging excessively from the reference policy \(\pi_{\text{ref}}\).
Moreover, a clipping term is applied to avoid overly large gradient steps, ensuring stable training.
The resulting objective is formulated as:
\begin{equation}
\begin{aligned}
\max_{\pi_\theta} \; \mathbb{E}_{\mathcal{O} \sim \pi_{\theta_{\text{old}}}(p)} \bigg[
& \min \left(
    \operatorname{clip} \left(
        \frac{\pi_\theta(o_i)}{\pi_{\theta_{\text{old}}}(o_i)},\;
        1 - \epsilon,\; 1 + \epsilon
    \right) A_i,\;
\right. \\
& \left. \qquad \frac{\pi_\theta(o_i)}{\pi_{\theta_{\text{old}}}(o_i)} A_i
\right)
- \beta\; \mathrm{D}_{\mathrm{KL}} \left( \pi_\theta \;\|\; \pi_{\text{ref}} \right)
\bigg]
\end{aligned} \label{eq:evigrpo}.
\end{equation}

\begin{figure}[!t]
\centering
\includegraphics[width=\linewidth]{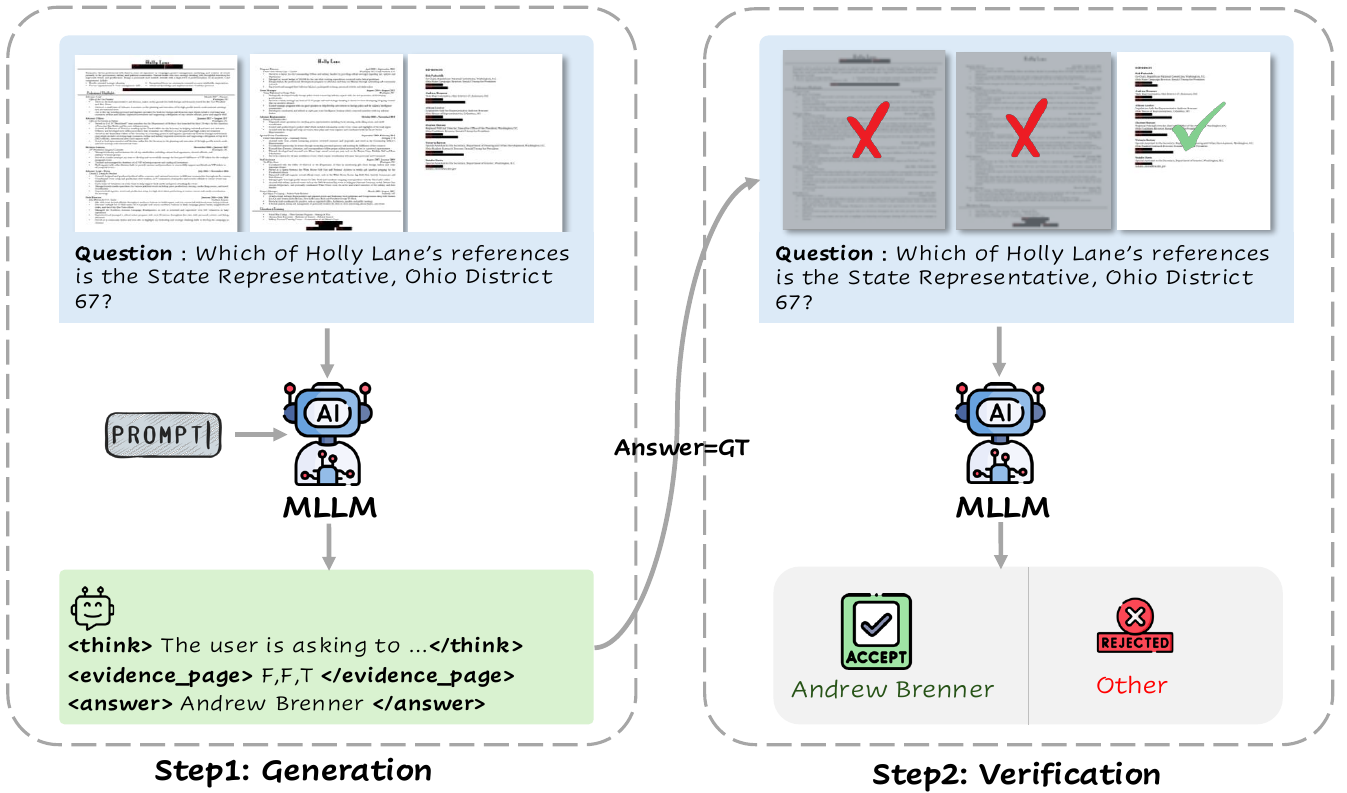}
\caption{The two-stage data annotation pipeline consists of a data generation process and an annotation verification process, designed to ensure the quality of the annotations.}
\label{fig:generate-reasongin-data}
\end{figure}

\subsection{Data Construction}
High-quality training data is critical for improving the multi-page document understanding capabilities of MLLMs.
However, most existing open-source datasets are not directly compatible with our EviGRPO framework due to mismatches in input-output formats and reward structures.
To address this, we design a new data construction pipeline and introduce \TrainData, the training dataset used in our study. To further ensure comprehensive evaluation across diverse multi-page document scenarios, we construct a new evaluation dataset, \AnnotationData, using the same annotation pipeline.

\subsubsection{Data Annotation Pipeline}
As illustrated in Figure~\ref{fig:generate-reasongin-data}, we develop a rigorous two-stage data annotation pipeline.
In the first stage, given an input and a task-specific prompt, we employ the Gemini 2.5 Flash model~\cite{comanici2025gemini} to generate a target output.
The sample advances to the next stage only if its predicted answer is consistent with the ground truth.
In the second stage, the same MLLM is prompted again with the annotated content in order to verify its accuracy.
The annotation is retained only if the model's output once again aligns with the ground truth under this controlled setting.

\begin{table}[!t]
\centering
\small
\scalebox{0.9}{
\begin{tabular}{@{}llcl@{}}
\toprule
\textbf{Category} & \textbf{Dataset} & \textbf{Images} & \textbf{\#Samples} \\
\midrule
\multirow{13}{*}{Single}
 & DocVQA             & 1       & 100  \\
 & InfographicVQA             & 1       & 100  \\
 & ChartQA                  & 1       & 100  \\
 & DeepForm                 & 1       & 100  \\
 & DVQA                      & 1       & 100  \\
 & FigureQA                  & 1       & 100  \\
 & KleisterCharity& 1       & 100  \\
 & OCRVQA                     & 1       & 100  \\
 & TabFact                   & 1       & 100  \\
 & TextCaps                  & 1       & 100  \\
 & TextVQA                  & 1       & 100  \\
 & VisualMRC                & 1       & 100  \\
 & WikiTableQuestions& 1       & 100  \\
\cmidrule{2-4}
 & \textbf{Total}      &       & \textbf{1300} \\
\midrule
\multirow{6}{*}{Multi}
 & DUDE              & 1--21   & 1000 \\
 & MP-DocVQA          & 1--36   & 500  \\
 & TATDoc            & 1--3    & 500  \\
 & SlideVQ           & 20      & 500  \\
 & Multihiertt        & 3--7    & 500  \\
 & \(\text{\AnnotationData}_\text{train}\)                 & 1--29   & 500  \\
\cmidrule{2-4}
 & \textbf{Total}       &       & \textbf{3500} \\
\bottomrule
\end{tabular}
}
\caption{Statistics of each dataset used for training.}
\label{tab:dataset_statis}
\end{table}

\begin{table*}[!t]
\centering
\small
\begin{tabular}{l |c| c  c c c c c c c c}
\toprule
\textbf{Model} & \textbf{Parameter} 
  & DocVQA & InfoVQA    & WTQ     & TabFact  
   & TextVQA    & VisualMRC \\
\midrule
UReader & 7B & 65.4 & 42.2 & 29.4 & 67.6 & 57.6 & 221.7 \\
TextMonkey & 9B & 73.0 & 28.6 & 31.9 & - & 65.9 & - \\
DocOwl-1.5-Chat & 8B & 82.2 & 50.7 & 40.6 & \textbf{80.2} & 68.6 & 246.4 \\
DocOwl-2 & 8B & 80.7 & 46.4 & 36.5 & 78.2 & 66.7 & 217.4 \\
Qwen2.5-VL-Instruct & 7B & 95.1 & 82.1  & 62.1 & 78.0 & \textbf{84.9} & \textbf{277.1}  \\
\midrule
\textbf{\approach}                    & \textbf{7B }
  & \textbf{95.1 }& \textbf{82.6 }
   & \textbf{63.1} & 79.6 
  & 81.0 & 251.6\\
\bottomrule
\end{tabular}

\caption{Performance comparison of 6 common single-page document benchmarks. InfoVQA is the abbreviation of the InfographicVQA dataset, and WTQ is the abbreviation of the WikiTableQuestions dataset. The best results are highlighted in \textbf{blod}.}
\label{tab:main-result-single}
\end{table*}

\begin{table*}[t]
\centering
\small
\scalebox{0.88}{
\begin{tabular}{l|c|ccccccc|c}
\toprule
\textbf{Models} & \textbf{Parameter} & MP-DocVQA & DUDE & SlideVQA & MultiChartQA & MultiHiertt & TATDoc & \AnnotationData & \textbf{Avg.} \\
\midrule

LLaVA-NeXT-Inter & 7B & 39.38 & 25.35 & 29.19 & 28.31 & 9.31 & 10.63 & 5.48 & 21.09  \\
LEOPARD-Idefics2 & 8B & 66.06 & 40.74 & 34.93 & 18.03 & 10.09 & 2.85 & 14.88 & 26.80 \\
mPlug-DocOwl2 & 8B & 67.98 & 31.25 & 29.55 & 4.85 & 8.08 & 22.16 & 7.12  & 24.43\\
LLaVA-oneVison & 7B & 49.38 & 31.43 & 45.94 & 32.73 & 10.53 & 15.42 & 10.38 & 27.97 \\


InternVL3-Instruct & 38B & 75.72 & 45.88 & 65.01 & 61.13 & 15.95 & 40.05 & 19.70 & 46.21 \\

Qwen2.5-VL-Instruct & 32B & 84.79 &  51.14 & 68.67 & 30.10 & 21.25 & 50.38 & 31.28 & 46.80 \\

Qwen2.5-VL-Instruct & 7B & 87.39 &  52.83 & 70.33 & 53.10 & 19.60 & 54.18 & 29.57 & 52.43 \\
\midrule
\textbf{\approach} & \textbf{7B} & \textbf{87.45} & \textbf{54.39} & \textbf{71.96} & \textbf{62.41} & \textbf{33.88} & \textbf{64.80} & \textbf{40.65} & \textbf{59.36} \\
\bottomrule
\end{tabular}
}
\caption{Performance comparison on 7 text-rich multi-page image datasets. All models are evaluated using ANLS metric. The best results are highlighted in \textbf{blod}.}
\label{tab:main-result}
\end{table*}

\subsubsection{\TrainData for Training}
As shown in Table~\ref{tab:dataset_statis}, based on the proposed annotation pipeline, we presents the statistics of our annotated dataset.
The dataset includes both single-page and multi-page document samples.
Specifically, for single-page documents, we curate 13 widely-used datasets, including DocVQA~\cite{mathew2021docvqa}, InfoGraphicsVQA~\cite{mathew2022infographicvqa}, ChartQA~\cite{masry2022chartqa}, DeepForm~\cite{svetlichnaya2020deepform}, DVQA~\cite{kafle2018dvqa}, FigureQA~\cite{kahou2017figureqa}, KleisterCharity~\cite{stanislawek2021kleister}, OCRVQA~\cite{mishra2019ocr}, TabFact~\cite{chen2019tabfact}, TextCaps~\cite{sidorov2020textcaps}, TextVQA~\cite{singh2019towards}, VisualMRC~\cite{tanaka2021visualmrc}, and WikiTableQuestions~\cite{pasupat2015compositional}.
These datasets cover a diverse range of document types.
For instance, DocVQA and DVQA contain scanned pages and visual charts, while DeepForm, TabFact, and WikiTableQuestions focus on structured tabular formats.
This diversity enables the model to learn from various layouts and visual structures commonly encountered in real-world documents.
For the multi-page document domain, we integrate three widely used datasets: DUDE~\cite{van2023document}, MP-DocVQA~\cite{tito2023hierarchical}, and TATDoc~\cite{zhu2022towards}, which are specifically designed for document-level reasoning across multiple pages.
To further enhance visual diversity, we also include SlideVQA~\cite{tanaka2023slidevqa}, a dataset based on multi-slide presentations, and Multihiertt~\cite{zhao2022multihiertt}, which focuses on complex hierarchical visual structures in multi-page charts.
In addition, academic paper reading is a crucial task in the multi-page document domain.
To improve the model's capability in this area, we annotate a training subset, denoted as \(\text{\AnnotationData}_\text{train}\), from the DocMatrix dataset~\cite{laurenccon2024building}, following the same annotation details described in the next section.

Specifically, we adopt the system prompt shown in Table~\ref{system prompt 3 ours}. For each of the 13 single-page datasets, we annotate 100 samples. For the multi-page datasets, we annotated 500 samples each, except for DUDE, which received 1,000 annotations due to its higher complexity.
In total, this process yielded a reasoning-annotated dataset \TrainData consisting of 1.3k single-page and 3.5k multi-page document samples.

\subsubsection{\AnnotationData for Testing}
A key application of multi-page document understanding is the comprehension of scientific papers.
However, large-scale multi-page benchmarks are still limited.
To fill this gap, we curate a subset from the large-scale DocMatrix dataset~\cite{laurenccon2024building}, which contains scientific papers sourced from ArXiv. Unlike \TrainData, the annotation details for \AnnotationData differ in two key aspects. First, as shown in the supplementary material, the prompt used for annotation is specifically tailored to scientific papers. Second, in step 1, the input consists solely of LaTeX-formatted text extracted from DocMatrix, rather than image-based document representations, in order to improve the accuracy of the generated annotations.  This enriched textual input guides the model in generating QA pairs across seven categories: factual, reasoning, comparison, summary, procedural, motivation, and result questions.
In Step 2, the input switches from LaTeX text to the visual format of the full paper, and the question is taken from the QA pair annotated in Step 1.
The data is retained only if the model's answer matches the previously annotated answer.
Following this process, we curate a new evaluation dataset, \AnnotationData, comprising 8.6k high-quality multi-page QA samples.

\subsection{Training Recipe}
We initialize our training with the Qwen2.5-VL-Instruct model~\cite{bai2025qwen2.5vl} for two main reasons.
First, large-scale chain-of-thought training data for multi-page document understanding is extremely scarce, and collecting such annotations is highly resource-intensive.
Second, this instruction-tuned model already exhibits a moderate level of reasoning ability, making it a practical alternative to the conventional ``cold start'' phase in GRPO.
To further adapt the model to our task, we adopt a two-stage curriculum training strategy within the EviGRPO framework, as illustrated in Figure~\ref{fig:model-overview}.
In the first stage, we train the model for one epoch using only single-page data. This step activates its latent reasoning capability while aligning its outputs with the expected answer format.
In the second stage, we continue training on multi-page data for another epoch to enhance the model's ability to reason over longer contexts and across multiple pages.

\begin{figure*}[!t]
\centering
\includegraphics[width=\textwidth]{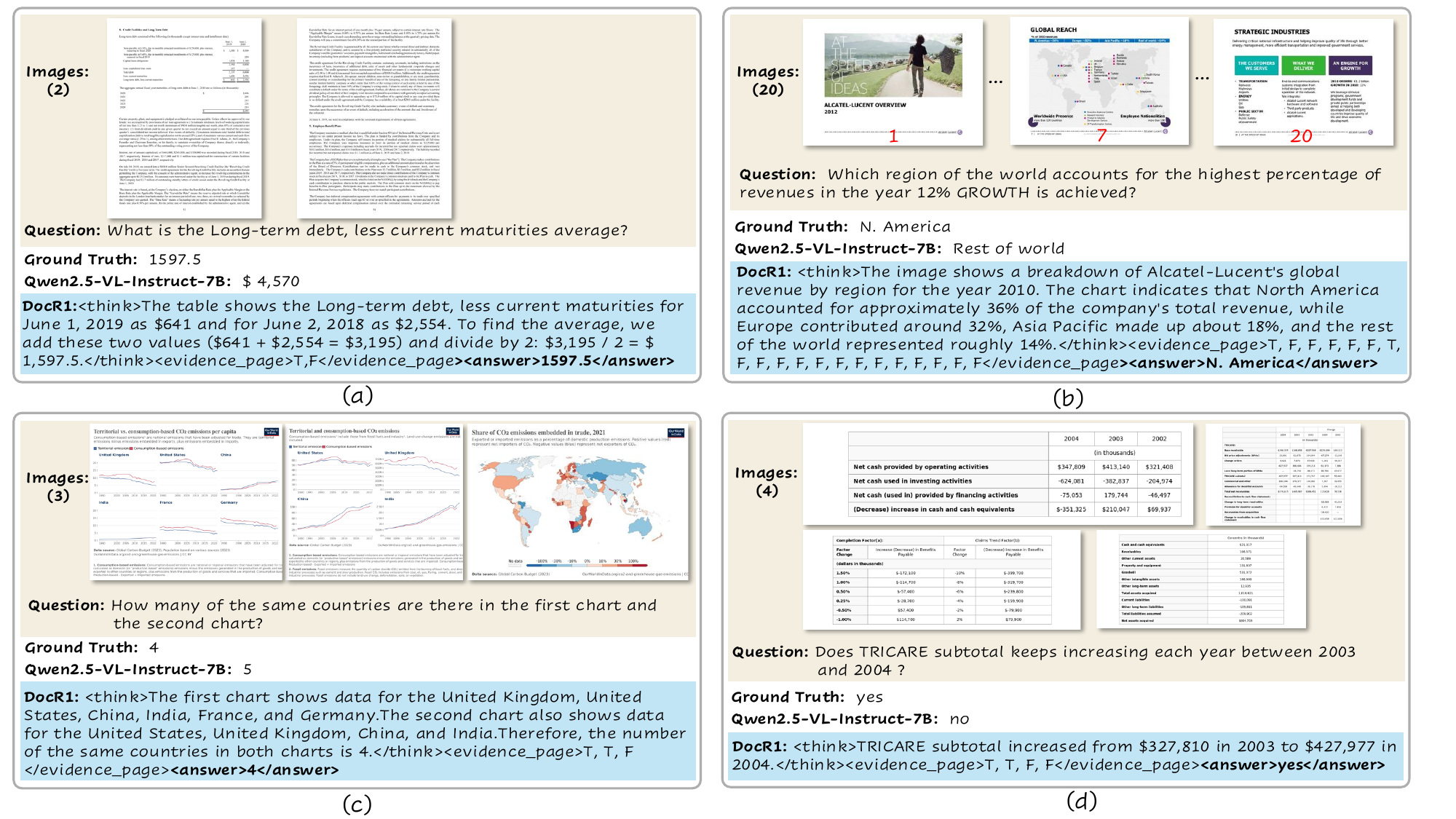}
\caption{Examples on multi-page document QA task. It can be seen that \approach can not only answer the questions correctly, but also provide relatively accurate evidence pages.}
\label{fig:case-study}
\end{figure*}

\section{Experiments}

\subsubsection{Implementation Details}
We conduct our experiments using 8 NVIDIA A100 GPUs.
During training, we set the batch size to 16, generate $G = 8$ candidate completions per sample, and use the KL penalty coefficient \(\beta = 0.04\). 
To improve computational efficiency, we constrain the input image resolution to a maximum of \(1024 \times 28 \times 28\) during both training and evaluation.

\subsubsection{Benchmarks} 
We evaluate our approach using both single-page and multi-page benchmarks.
For single-page evaluation, we consider six widely used datasets: DocVQA~\cite{mathew2021docvqa}, InfographicVQA~\cite{mathew2022infographicvqa}, WikiTableQuestions~\cite{pasupat2015compositional}, TabFact~\cite{chen2019tabfact}, TextVQA~\cite{singh2019towards}, and VisualMRC~\cite{tanaka2021visualmrc}.
For multi-page document understanding, we assess our method on six publicly available benchmarks: DUDE~\cite{van2023document}, MP-DocVQA~\cite{tito2023hierarchical}, TATDoc~\cite{zhu2022towards}, Multi-ChartQA~\cite{zhu2024multichartqa}, MultiHiertt~\cite{zhao2022multihiertt}, and SlideVQA~\cite{tanaka2023slidevqa}, as well as our newly annotated benchmark, \AnnotationData, to comprehensively evaluate its effectiveness in processing complex, multi-page inputs.
Detailed dataset statistics are provided in the supplementary material.

\subsection{Main Result}
\subsubsection{Quantitative Results}
As shown in Table~\ref{tab:main-result-single}, \approach demonstrates strong performance across six widely used single-page document benchmarks.
It performs comparably to Qwen2.5-VL-Instruct, with particularly notable improvements on WikiTableQuestions and TabFact, where it outperforms the baseline by 1.0 and 1.6 points, respectively.
In Table~\ref{tab:main-result}, our method achieves state-of-the-art performance across all multi-page document understanding benchmarks.
Remarkably, despite using significantly fewer parameters, \approach consistently surpasses much larger models such as Qwen2.5-VL-Instruct (32B) and InternVL-3-Instruct (38B).
It attains an average score of 59.36, representing a absolute improvement of 6.93 over the baseline.
Notably, our method yields improvements of over 10 points on four particularly challenging benchmarks—MultiChartQA, MultiHiertt, TATDoc, and \AnnotationData.
These results highlight the effectiveness of the proposed EviGRPO framework and training strategy. With only 4.8k annotated samples, \approach significantly improves multi-page reasoning capabilities while preserving, and in some cases slightly enhancing performance on single-page benchmarks. This demonstrates not only high data efficiency but also strong generalization from coarse-grained page-level retrieval to fine-grained semantic understanding.

\subsubsection{Qualitative Results}
As illustrated in Figure~\ref{fig:case-study}, we provide visual comparisons between \approach and the baseline model across different benchmarks.
In Figure~\ref{fig:case-study}(a), given the question ``What is the Long-term debt, less current maturities average?'', the baseline model fails to locate the relevant content and outputs an incorrect answer.
In contrast, \approach accurately identifies the first page as the sole evidence source, extracts the correct financial figure, and generates a coherent reasoning trace that leads to the correct answer.
Similar trends are observed in the remaining three examples.
Our evidence-first strategy significantly enhances the model's coarse-to-fine reasoning ability, allowing it to concentrate on a small set of relevant pages.
Even in challenging cases such as Figure~\ref{fig:case-study}(b), which involves 20 pages, \approach effectively identifies page 7 as the critical source of evidence and constructs an accurate reasoning path grounded in its content.
These results demonstrate the model's strong capability in localizing sparse information and performing structured reasoning in complex multi-page scenarios.

\begin{table*}[t]
\centering
\small
\scalebox{0.92}{
    \begin{tabular}{lccccccc}
    \toprule
    \textbf{Method} & MP-DocVQA & DUDE & SlideVQA & MultiChartQA & MultiHiertt & TATDoc & \AnnotationData \\
    \midrule
    \multicolumn{4}{l}{\textbf{Qwen2.5-VL-Instruct }} \\
    \quad Baseline  & 87.39 &  52.83 & 70.33 & 53.10 & 19.60 & 54.18 & 29.57  \\
    \quad SFT(w/ mixdata)           & 87.25(-0.14)  & 52.72(-0.11)  &71.30 \textbf{(+0.97)} & 51.34(-1.76) & 20.76\textbf{(+1.16)} & 55.61\textbf{(+1.43)} & 37.93\textbf{(+8.36)}\\
    \quad GRPO(w/ mixdata)           & 87.49\textbf{(+0.10)}  & 53.14\textbf{(+0.31)}  &70.92 \textbf{(+0.59)} & 54.26\textbf{(+1.16)} & 29.02\textbf{(+9.42)} & 60.03\textbf{(+5.85)} & 38.11\textbf{(+8.54)}\\
    \midrule
    \multicolumn{4}{l}{\textbf{EviGRPO }} \\
    \quad w/ single           & 76.95(-10.44) & 46.45(-6.38) & 49.11(-21.22) & 49.09(-4.01) & 25.26\textbf{(+5.66)} & 56.11\textbf{(+1.93)} &27.04(-2.53) \\
    \quad w/ multi            & 86.82(-0.57)  & 52.84\textbf{(+0.01)} & 70.71\textbf{(+0.38)} &60.82\textbf{(+7.72)} & 28.33\textbf{(+8.73)} & 64.15\textbf{(+9.97)} & 40.03\textbf{(+10.46)} \\
    \quad w/ mixdata          & 86.85(-0.54) & 52.93\textbf{(+0.10)} & 71.24\textbf{(+0.91)} & 61.54\textbf{(+8.44)} & 29.49\textbf{(+9.89)} & 64.51\textbf{(+10.33)} & 39.93\textbf{(+10.36)}\\
    \midrule
    \textbf{Page Selection} \\ 
    \quad PSF-1& 87.06(-0.33) & 52.57(-0.26) & 69.84(-0.49) & 56.27(\textbf{+3.17}) &32.94\textbf{(+13.34)} & 60.93\textbf{(+6.75)}  & 40.36\textbf{(+10.69)} \\
    \quad PSF-2& 86.51(-0.88) & 52.45(-0.38) & 71.22\textbf{(+0.89)} & 61.64\textbf{(+8.54)} & 32.58\textbf{(+12.98)} & 61.84\textbf{(+7.66)} & 41.02\textbf{(+11.45)}\\
    \midrule
    \textbf{Ours} & 87.45\textbf{(+0.06)} & 54.39\textbf{(+1.56)} & 71.96\textbf{(+1.63)} & 62.41\textbf{(+9.31)} & 33.88\textbf{(+14.28)} & 64.80\textbf{(+10.62)} & 40.65\textbf{(+11.08)} \\
    \bottomrule
    \end{tabular}
}
\caption{Ablation experiments on training paradigms, data composition, training strategies, and page selection formats (PSF).To ensure consistency, all EviGRPO experiments adopt the same page selection format as Ours.}
\label{tab:ablation_study_train}
\end{table*} 

\subsection{Ablation Studies}

\subsubsection{Training Paradigm: SFT vs. GRPO vs. EviGRPO}
To assess the impact of different training paradigms, we compare Supervised Fine-Tuning (SFT), standard GRPO, and our proposed EviGRPO using the same set of mixed single-page and multi-page training data (denoted as mixdata).
As shown in Table~\ref{tab:ablation_study_train}, SFT achieves only modest improvements on some benchmarks, primarily due to its limited data efficiency. Notably, \AnnotationData is the only benchmark where SFT shows a significant gain, likely due to the model’s exposure to previously unseen data domains. In contrast, both GRPO and EviGRPO demonstrate substantial performance gains on nearly all benchmarks, underscoring the effectiveness of RL in enhancing reasoning capabilities by leveraging limited supervision more effectively. Furthermore, EviGRPO generally outperforms standard GRPO, demonstrating that its coarse-to-fine reasoning mechanism enables the model to identify relevant evidence pages prior to answer generation, thereby leading to more accurate predictions.

\subsubsection{Data Composition: Single vs. Multi-page}
To assess the effectiveness of different training subsets, we compare EviGRPO trained with only single-page data or multi-page data. As shown in Table~\ref{tab:ablation_study_train}, training solely on single-page data severely compromises the model’s multi-page reasoning capabilities, as such data lacks the need for evidence page identification—leading to significant performance drops on most multi-page tasks. In contrast, training with only multi-page data yields performance gains across many benchmarks, but the improvements are consistently smaller than those achieved when both data types are combined. These results underscore the importance of leveraging both single-page and multi-page data to fully unlock the model’s multi-page reasoning potential.

\subsubsection{Training Strategy: Mixed vs. Curriculum}
To explore different strategies for combining single-page and multi-page data, we further compare our two-stage curriculum training approach with a baseline that trains on mixed data simultaneously. As shown in Table~\ref{tab:ablation_study_train}, although the mixed strategy also yields improvements across most benchmarks, our curriculum-based method consistently delivers greater and more generalized gains. This can be attributed to the ``cold-start'' effect of single-page training, which helps the model become familiar with the required output format and reasoning style before tackling more complex multi-page tasks.

\subsubsection{Page Selection Format Variants}
In addition, we compare three page selection formats designed to guide evidence identification in multi-page document reasoning. PSF-1 prompts the model to directly output the indices of relevant pages (e.g., “1, 3”). PSF-2 requires the model to assign a binary label ``T'' (True) or ``F'' (False), with the total number of images explicitly provided. Ours differs from PSF-2 by omitting the image count.
As shown in Table~\ref{tab:ablation_study_train}, PSF-1 encourages sparse selection, often leading to the omission of relevant pages due to the lack of enforced per-image decisions. PSF-2 alleviates this limitation by requiring exhaustive labeling; however, the known number of input images enables the model to heuristically align its outputs without performing genuine reasoning. In contrast, our method strengthens the reasoning requirement by omitting the image count, compelling the model to first infer the number of input pages before assigning labels accordingly.
Detailed specifications of PSF-1 and PSF-2 are provided in the supplementary materials.

\subsubsection{Evidence Page Selection Accuracy}
To quantitatively evaluate the accuracy of the evidence page selection mechanism, we compute the recall of predicted evidence pages across three multi-page document understanding benchmarks.
As shown in Table~\ref{tab:evidence_recall}, \approach achieves an average recall improvement of 38.77 points over the baseline, demonstrating its superior ability to identify ground-truth evidence pages.
These results underscore the reliability of the coarse-to-fine reasoning strategy employed in our framework.

\begin{table}[!t]
\centering
\begin{tabular}{lcc}
\toprule
\textbf{Dataset} & \textbf{Qwen2.5-VL-Instruct }& \textbf{\approach} \\
\midrule
MP-DocVQA   & 46.13 & 91.69\textbf{(+45.56)} \\
DUDE        & 56.34 & 85.33\textbf{(+28.99)} \\
MultiHiertt & 55.62 & 97.38\textbf{(+41.76)} \\
\midrule
\textbf{Avg} & 52.70 & 91.47\textbf{(+38.77)} \\
\bottomrule
\end{tabular}
\caption{Evidence page recall on three multi-page document benchmarks. Recall is used as the evaluation metric to measure whether the ground truth evidence pages are successfully retrieved by the model.}
\label{tab:evidence_recall}
\end{table}


\section{Conclusion}
In this work, we propose \approach, an MLLM specifically designed for multi-page document understanding, trained under our newly introduced RL framework, EviGRPO. By incorporating evidence-aware rewards, EviGRPO encourages a human-like coarse-to-fine reasoning process, enabling the model to effectively retrieve and reason over relevant content. Supported by a two-stage annotation pipeline and a curriculum training strategy, our approach enables efficient learning under limited supervision. Extensive experiments across multiple benchmarks demonstrate that \approach achieves state-of-the-art performance on multi-page tasks while maintaining strong capabilities on single-page inputs. This work highlights the potential of RL to advance document-level reasoning in MLLMs.

\bibliography{aaai26}

@article{li2025videochat,
  title={{Videochat-R1}: Enhancing spatio-temporal perception via reinforcement fine-tuning},
  author={Li, Xinhao and Yan, Ziang and Meng, Desen and Dong, Lu and Zeng, Xiangyu and He, Yinan and Wang, Yali and Qiao, Yu and Wang, Yi and Wang, Limin},
  journal={arXiv preprint arXiv:2504.06958},
  year={2025}
}

@article{huang2025vision,
  title={{Vision-R1}: Incentivizing reasoning capability in multimodal large language models},
  author={Huang, Wenxuan and Jia, Bohan and Zhai, Zijie and Cao, Shaosheng and Ye, Zheyu and Zhao, Fei and Xu, Zhe and Hu, Yao and Lin, Shaohui},
  journal={arXiv preprint arXiv:2503.06749},
  year={2025}
}

@article{feng2025video,
  title={{Video-R1}: Reinforcing video reasoning in mllms},
  author={Feng, Kaituo and Gong, Kaixiong and Li, Bohao and Guo, Zonghao and Wang, Yibing and Peng, Tianshuo and Wu, Junfei and Zhang, Xiaoying and Wang, Benyou and Yue, Xiangyu},
  journal={arXiv preprint arXiv:2503.21776},
  year={2025}
}

@article{liu2025visual,
  title={{Visual-RFT}: Visual reinforcement fine-tuning},
  author={Liu, Ziyu and Sun, Zeyi and Zang, Yuhang and Dong, Xiaoyi and Cao, Yuhang and Duan, Haodong and Lin, Dahua and Wang, Jiaqi},
  journal={arXiv preprint arXiv:2503.01785},
  year={2025}
}

@article{ye2023ureader,
  title={{UReader}: Universal ocr-free visually-situated language understanding with multimodal large language model},
  author={Ye, Jiabo and Hu, Anwen and Xu, Haiyang and Ye, Qinghao and Yan, Ming and Xu, Guohai and Li, Chenliang and Tian, Junfeng and Qian, Qi and Zhang, Ji and others},
  journal={arXiv preprint arXiv:2310.05126},
  year={2023}
}

@article{liu2024textmonkey,
  title={{TextMonkey}: An ocr-free large multimodal model for understanding document},
  author={Liu, Yuliang and Yang, Biao and Liu, Qiang and Li, Zhang and Ma, Zhiyin and Zhang, Shuo and Bai, Xiang},
  journal={arXiv preprint arXiv:2403.04473},
  year={2024}
}

@article{ye2023mplug,
  title={{mplug-docowl}: Modularized multimodal large language model for document understanding},
  author={Ye, Jiabo and Hu, Anwen and Xu, Haiyang and Ye, Qinghao and Yan, Ming and Dan, Yuhao and Zhao, Chenlin and Xu, Guohai and Li, Chenliang and Tian, Junfeng and others},
  journal={arXiv preprint arXiv:2307.02499},
  year={2023}
}

@article{hu2024mplug1.5,
  title={{mplug-docowl 1.5}: Unified structure learning for ocr-free document understanding},
  author={Hu, Anwen and Xu, Haiyang and Ye, Jiabo and Yan, Ming and Zhang, Liang and Zhang, Bo and Li, Chen and Zhang, Ji and Jin, Qin and Huang, Fei and others},
  journal={arXiv preprint arXiv:2403.12895},
  year={2024}
}

@article{hu2024mplug2,
  title={{mplug-docowl2}: High-resolution compressing for ocr-free multi-page document understanding},
  author={Hu, Anwen and Xu, Haiyang and Zhang, Liang and Ye, Jiabo and Yan, Ming and Zhang, Ji and Jin, Qin and Huang, Fei and Zhou, Jingren},
  journal={arXiv preprint arXiv:2409.03420},
  year={2024}
}

@article{mo2025doc,
  title={{Doc-CoB}: Enhancing Multi-Modal Document Understanding with Visual Chain-of-Boxes Reasoning},
  author={Mo, Ye and Shao, Zirui and Ye, Kai and Mao, Xianwei and Zhang, Bo and Xing, Hangdi and Ye, Peng and Huang, Gang and Chen, Kehan and Huan, Zhou and others},
  journal={arXiv preprint arXiv:2505.18603},
  year={2025}
}

@article{zhou2024doge,
  title={{Doge}: Towards versatile visual document grounding and referring},
  author={Zhou, Yinan and Chen, Yuxin and Lin, Haokun and Yang, Shuyu and Zhu, Li and Qi, Zhongang and Ma, Chen and Shan, Ying},
  journal={arXiv preprint arXiv:2411.17125},
  year={2024}
}

@inproceedings{van2023document,
  title={Document understanding dataset and evaluation (dude)},
  author={Van Landeghem, Jordy and Tito, Rub{\`e}n and Borchmann, {\L}ukasz and Pietruszka, Micha{\l} and Joziak, Pawel and Powalski, Rafal and Jurkiewicz, Dawid and Coustaty, Micka{\"e}l and Anckaert, Bertrand and Valveny, Ernest and others},
  booktitle={Proceedings of the IEEE International Conference on Computer Vision},
  pages={19528--19540},
  year={2023}
}

@article{tito2023hierarchical,
  title={Hierarchical multimodal transformers for multipage docvqa},
  author={Tito, Rub{\`e}n and Karatzas, Dimosthenis and Valveny, Ernest},
  journal={Pattern Recognition},
  volume={144},
  pages={109834},
  year={2023}
}

@inproceedings{zhu2022towards,
  title={Towards complex document understanding by discrete reasoning},
  author={Zhu, Fengbin and Lei, Wenqiang and Feng, Fuli and Wang, Chao and Zhang, Haozhou and Chua, Tat-Seng},
  booktitle={Proceedings of the ACM International Conference on Multimedia},
  pages={4857--4866},
  year={2022}
}

@inproceedings{tanaka2023slidevqa,
  title={{Slidevqa}: A dataset for document visual question answering on multiple images},
  author={Tanaka, Ryota and Nishida, Kyosuke and Nishida, Kosuke and Hasegawa, Taku and Saito, Itsumi and Saito, Kuniko},
  booktitle={Proceedings of the AAAI Conference on Artificial Intelligence},
  pages={13636--13645},
  year={2023}
}

@article{zhao2022multihiertt,
  title={{MultiHiertt}: Numerical reasoning over multi hierarchical tabular and textual data},
  author={Zhao, Yilun and Li, Yunxiang and Li, Chenying and Zhang, Rui},
  journal={arXiv preprint arXiv:2206.01347},
  year={2022}
}

@article{zhu2024multichartqa,
  title={{MultiChartQA}: Benchmarking Vision-Language Models on Multi-Chart Problems},
  author={Zhu, Zifeng and Jia, Mengzhao and Zhang, Zhihan and Li, Lang and Jiang, Meng},
  journal={arXiv preprint arXiv:2410.14179},
  year={2024}
}

@inproceedings{mathew2021docvqa,
  title={{Docvqa}: A dataset for vqa on document images},
  author={Mathew, Minesh and Karatzas, Dimosthenis and Jawahar, CV},
  booktitle={Proceedings of the IEEE Winter Conference on Applications of Computer Vision},
  pages={2200--2209},
  year={2021}
}

@article{kahou2017figureqa,
  title={{FigureQA}: An annotated figure dataset for visual reasoning},
  author={Kahou, Samira Ebrahimi and Michalski, Vincent and Atkinson, Adam and K{\'a}d{\'a}r, {\'A}kos and Trischler, Adam and Bengio, Yoshua},
  journal={arXiv preprint arXiv:1710.07300},
  year={2017}
}

@inproceedings{mathew2022infographicvqa,
  title={Infographicvqa},
  author={Mathew, Minesh and Bagal, Viraj and Tito, Rub{\`e}n and Karatzas, Dimosthenis and Valveny, Ernest and Jawahar, CV},
  booktitle={Proceedings of the IEEE Winter Conference on Applications of Computer Vision},
  pages={1697--1706},
  year={2022}
}

@article{masry2022chartqa,
  title={{Chartqa}: A benchmark for question answering about charts with visual and logical reasoning},
  author={Masry, Ahmed and Long, Do Xuan and Tan, Jia Qing and Joty, Shafiq and Hoque, Enamul},
  journal={arXiv preprint arXiv:2203.10244},
  year={2022}
}

@article{svetlichnaya2020deepform,
  title={{DeepForm}: Understand structured documents at scale},
  author={Svetlichnaya, Stacey},
    journal = {arXiv preprint},
  year={2020}
}

@inproceedings{stanislawek2021kleister,
  title={{Kleister}: key information extraction datasets involving long documents with complex layouts},
  author={Stanis{\l}awek, Tomasz and Grali{\'n}ski, Filip and Wr{\'o}blewska, Anna and Lipi{\'n}ski, Dawid and Kaliska, Agnieszka and Rosalska, Paulina and Topolski, Bartosz and Biecek, Przemys{\l}aw},
  booktitle={Proceedings of the International Conference on Document Analysis and Recognition},
  pages={564--579},
  year={2021}
}

@article{pasupat2015compositional,
  title={Compositional semantic parsing on semi-structured tables},
  author={Pasupat, Panupong and Liang, Percy},
  journal={arXiv preprint arXiv:1508.00305},
  year={2015}
}

@inproceedings{tanaka2021visualmrc,
  title={{VisualMRC}: Machine reading comprehension on document images},
  author={Tanaka, Ryota and Nishida, Kyosuke and Yoshida, Sen},
  booktitle={Proceedings of the AAAI Conference on Artificial Intelligence},
  pages={13878--13888},
  year={2021}
}

@inproceedings{sidorov2020textcaps,
  title={{Textcaps}: a dataset for image captioning with reading comprehension},
  author={Sidorov, Oleksii and Hu, Ronghang and Rohrbach, Marcus and Singh, Amanpreet},
  booktitle={Proceedings of the European Conference on Computer Vision},
  pages={742--758},
  year={2020}
}

@inproceedings{kafle2018dvqa,
  title={{Dvqa}: Understanding data visualizations via question answering},
  author={Kafle, Kushal and Price, Brian and Cohen, Scott and Kanan, Christopher},
  booktitle={Proceedings of the IEEE Conference on Computer Vision and Pattern Recognition},
  pages={5648--5656},
  year={2018}
}

@article{chen2019tabfact,
  title={{Tabfact}: A large-scale dataset for table-based fact verification},
  author={Chen, Wenhu and Wang, Hongmin and Chen, Jianshu and Zhang, Yunkai and Wang, Hong and Li, Shiyang and Zhou, Xiyou and Wang, William Yang},
  journal={arXiv preprint arXiv:1909.02164},
  year={2019}
}

@inproceedings{mishra2019ocr,
  title={{Ocr-vqa}: Visual question answering by reading text in images},
  author={Mishra, Anand and Shekhar, Shashank and Singh, Ajeet Kumar and Chakraborty, Anirban},
  booktitle={Proceedings of the International Conference on Document Analysis and Recognition},
  pages={947--952},
  year={2019}
}

@inproceedings{singh2019towards,
  title={Towards vqa models that can read},
  author={Singh, Amanpreet and Natarajan, Vivek and Shah, Meet and Jiang, Yu and Chen, Xinlei and Batra, Dhruv and Parikh, Devi and Rohrbach, Marcus},
  booktitle={Proceedings of the IEEE Conference on Computer Vision and Pattern Recognition},
  pages={8317--8326},
  year={2019}
}

@article{jia2024leopard,
  title={{Leopard}: A vision language model for text-rich multi-image tasks},
  author={Jia, Mengzhao and Yu, Wenhao and Ma, Kaixin and Fang, Tianqing and Zhang, Zhihan and Ouyang, Siru and Zhang, Hongming and Jiang, Meng and Yu, Dong},
  journal={arXiv preprint arXiv:2410.01744},
  year={2024}
}

@article{jaech2024openai,
  title={Openai o1 system card},
  author={Jaech, Aaron and Kalai, Adam and Lerer, Adam and Richardson, Adam and El-Kishky, Ahmed and Low, Aiden and Helyar, Alec and Madry, Aleksander and Beutel, Alex and Carney, Alex and others},
  journal={arXiv preprint arXiv:2412.16720},
  year={2024}
}

@article{guo2025deepseek,
  title={{Deepseek-r1}: Incentivizing reasoning capability in llms via reinforcement learning},
  author={Guo, Daya and Yang, Dejian and Zhang, Haowei and Song, Junxiao and Zhang, Ruoyu and Xu, Runxin and Zhu, Qihao and Ma, Shirong and Wang, Peiyi and Bi, Xiao and others},
  journal={arXiv preprint arXiv:2501.12948},
  year={2025}
}

@article{wang2023towards,
  title={Towards improving document understanding: An exploration on text-grounding via mllms},
  author={Wang, Yonghui and Zhou, Wengang and Feng, Hao and Zhou, Keyi and Li, Houqiang},
  journal={arXiv preprint arXiv:2311.13194},
  year={2023}
}

@article{shao2024deepseekmath,
  title={{Deepseekmath}: Pushing the limits of mathematical reasoning in open language models},
  author={Shao, Zhihong and Wang, Peiyi and Zhu, Qihao and Xu, Runxin and Song, Junxiao and Bi, Xiao and Zhang, Haowei and Zhang, Mingchuan and Li, YK and Wu, Yang and others},
  journal={arXiv preprint arXiv:2402.03300},
  year={2024}
}

@article{laurenccon2024building,
  title={Building and better understanding vision-language models: insights and future directions},
  author={Lauren{\c{c}}on, Hugo and Marafioti, Andr{\'e}s and Sanh, Victor and Tronchon, L{\'e}o},
  journal={arXiv preprint arXiv:2408.12637},
  year={2024}
}

@article{comanici2025gemini,
  title={{Gemini 2.5}: Pushing the Frontier with Advanced Reasoning, Multimodality, Long Context, and Next Generation Agentic Capabilities},
  author={Comanici, Gheorghe and Bieber, Eric and Schaekermann, Mike and Pasupat, Ice and Sachdeva, Noveen and Dhillon, Inderjit and Blistein, Marcel and Ram, Ori and Zhang, Dan and Rosen, Evan and others},
  journal={arXiv preprint arXiv:2507.06261},
  year={2025}
}

@article{bai2025qwen2.5vl,
  title={Qwen2. 5-vl technical report},
  author={Bai, Shuai and Chen, Keqin and Liu, Xuejing and Wang, Jialin and Ge, Wenbin and Song, Sibo and Dang, Kai and Wang, Peng and Wang, Shijie and Tang, Jun and others},
  journal={arXiv preprint arXiv:2502.13923},
  year={2025}
}

@inproceedings{zhao2024tabpedia,
  title={{Tabpedia:} Towards comprehensive visual table understanding with concept synergy},
  author={Zhao, Weichao and Feng, Hao and Liu, Qi and Tang, Jingqun and Wu, Binghong and Liao, Lei and Wei, Shu and Ye, Yongjie and Liu, Hao and Zhou, Wengang and others},
  booktitle={Proceedings of the Advances in Neural Information Processing Systems},
  pages={7185--7212},
  year={2024}
}

@article{feng2024docpedia,
  title={{Docpedia:} Unleashing the power of large multimodal model in the frequency domain for versatile document understanding},
  author={Feng, Hao and Liu, Qi and Liu, Hao and Tang, Jingqun and Zhou, Wengang and Li, Houqiang and Huang, Can},
  journal={Science China Information Sciences},
  volume={67},
  pages={220106},
  year={2024},
  publisher={Springer}
}

@misc{dong2025spreadsheetllmencodingspreadsheetslarge,
      title={SpreadsheetLLM: Encoding Spreadsheets for Large Language Models}, 
      author={Haoyu Dong and Jianbo Zhao and Yuzhang Tian and Junyu Xiong and Shiyu Xia and Mengyu Zhou and Yun Lin and José Cambronero and Yeye He and Shi Han and Dongmei Zhang},
      year={2025},
      eprint={2407.09025},
      archivePrefix={arXiv},
      primaryClass={cs.AI},
      url={https://arxiv.org/abs/2407.09025}, 
}

@inproceedings{Xia_2024,
   title={Vision Language Models for Spreadsheet Understanding: Challenges and Opportunities},
   url={http://dx.doi.org/10.18653/v1/2024.alvr-1.10},
   DOI={10.18653/v1/2024.alvr-1.10},
   booktitle={Proceedings of the 3rd Workshop on Advances in Language and Vision Research (ALVR)},
   publisher={Association for Computational Linguistics},
   author={Xia, Shiyu and Xiong, Junyu and Dong, Haoyu and Zhao, Jianbo and Tian, Yuzhang and Zhou, Mengyu and He, Yeye and Han, Shi and Zhang, Dongmei},
   year={2024},
   pages={116–128} }
\appendix
\begin{table}[htbp]
\centering
\small
\scalebox{1}{
\begin{tabular}{@{}llcl@{}}
\toprule
\textbf{Category} & \textbf{Dataset} & \textbf{Images} & \textbf{\#Samples} \\
\midrule
\multirow{6}{*}{Single}
 & DocVQA              & 1       & 5,188  \\
 & InfographicVQA             & 1       & 3,288  \\

 & TabFact                  & 1       & 12,722  \\

 & TextVQA                   & 1       & 5,000  \\
 & VisualMRC                 & 1       & 6,729  \\
 & WikiTableQuestions         & 1       & 4,343  \\

\midrule
\multirow{7}{*}{Multi}
 & DUDE              & 1--20   & 11,402 \\
 & MP-DocVQA          & 1-20   & 5019  \\
 & TATDoc             & 1--2    & 1663  \\
 & SlideVQA           & 20      & 2,136  \\
  & MultiChartQA           & 2-3      & 2,000  \\
 & Multihiert      & 3--7    & 7,830  \\
 & \AnnotationData                 & 1--30   & 8,648  \\

\bottomrule
\end{tabular}
}
\caption{Statistics of various benchmark datasets used for evaluation.}
\label{tab:dataset_statis_benchmark}
\end{table}

\section{Evaluation Benchmarks}

To evaluate the effectiveness of our approach, we select six single-page and seven multi-page document understanding benchmarks. Notably, for datasets whose official test sets are not publicly accessible—TextVQA, DUDE, MP-DocVQA, and MultiHiertt—we follow prior works~\cite{jia2024leopard,zhou2024doge} and report results on their validation sets.

The number of images and samples in each dataset is summarized in Table~\ref{tab:dataset_statis_benchmark}.

\section{EviGRPO Training Pseudocode}
\label{app:pseudocode}

\begin{figure*}[htbp]
  \centering

  \includegraphics[width=\linewidth]{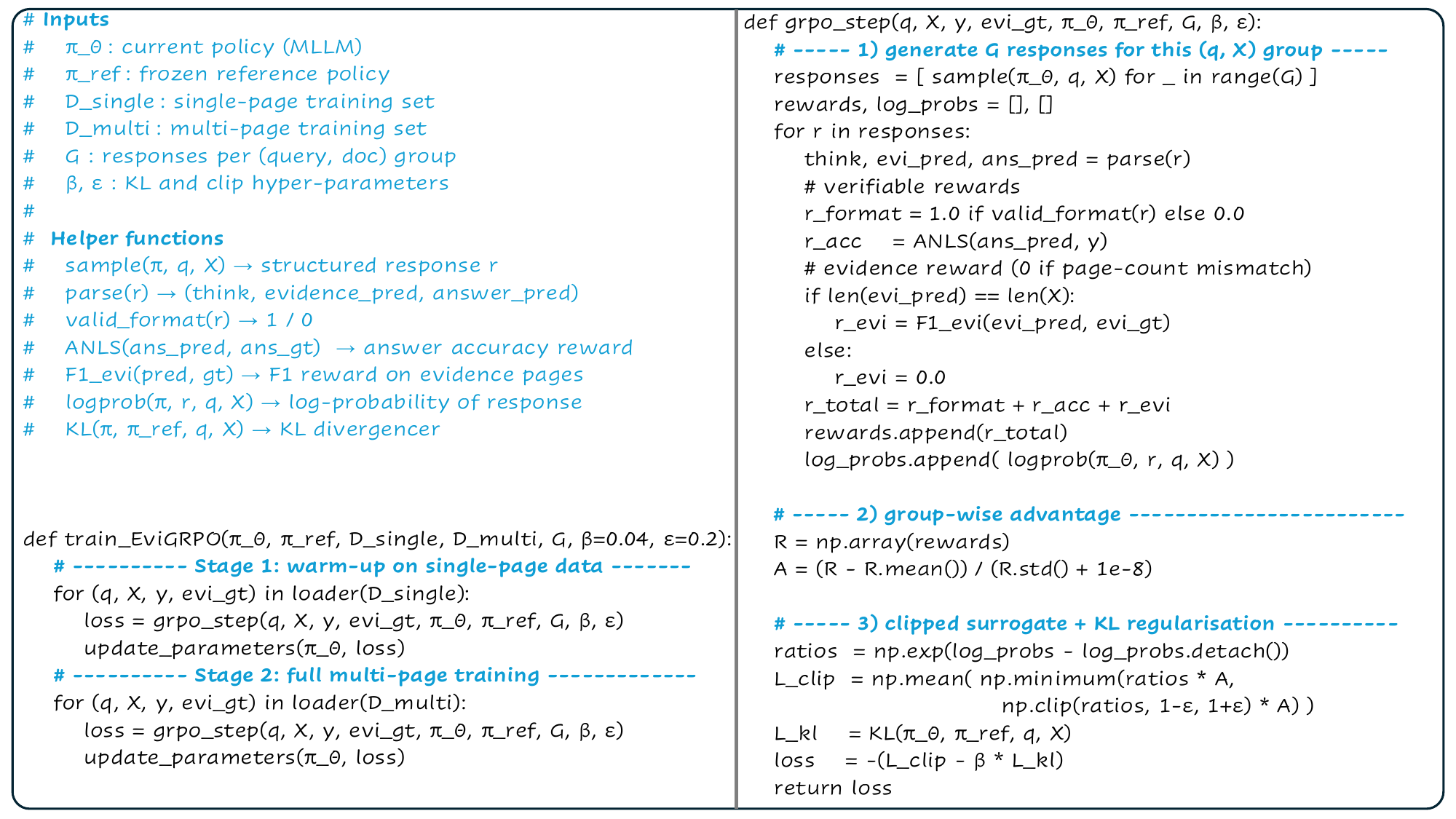}
  \caption{Numpy-style pseudocode of the EviGRPO training loop used for
           DocR1.}
  \label{fig:evi_pseudocode}
\end{figure*}

Figure~\ref{fig:evi_pseudocode} presents the pseudocode for training our model with EviGRPO. The training consists of two stages: a warm-up stage on single-page data and a full training stage on multi-page data. For each batch, the \texttt{grpo\_step} function generates $G$ responses for a given input triplet $(q, X)$ using the current policy $\pi_\theta$. Each generated response is parsed into its components—intermediate reasoning, predicted evidence pages, and answer—and is evaluated using three verifiable reward metrics: \texttt{valid\_format} for structural consistency, \texttt{ANLS} for answer accuracy, and \texttt{F1\_evi} for evidence selection. The total reward is the sum of these components. If the predicted evidence page count mismatches the ground truth, the evidence reward is skipped. Then, normalized group-wise advantages are computed using the mean and standard deviation of the total rewards. The policy is updated using a clipped surrogate loss, combined with a KL divergence term that penalizes deviation from a frozen reference policy $\pi_\text{ref}$. The final loss balances both terms with a weight $\beta$, and $\epsilon$ controls the clipping threshold.

\section{Prompt and Example of ArxivFullQA}
\begin{figure*}[t]
\centering
\includegraphics[width=\textwidth]{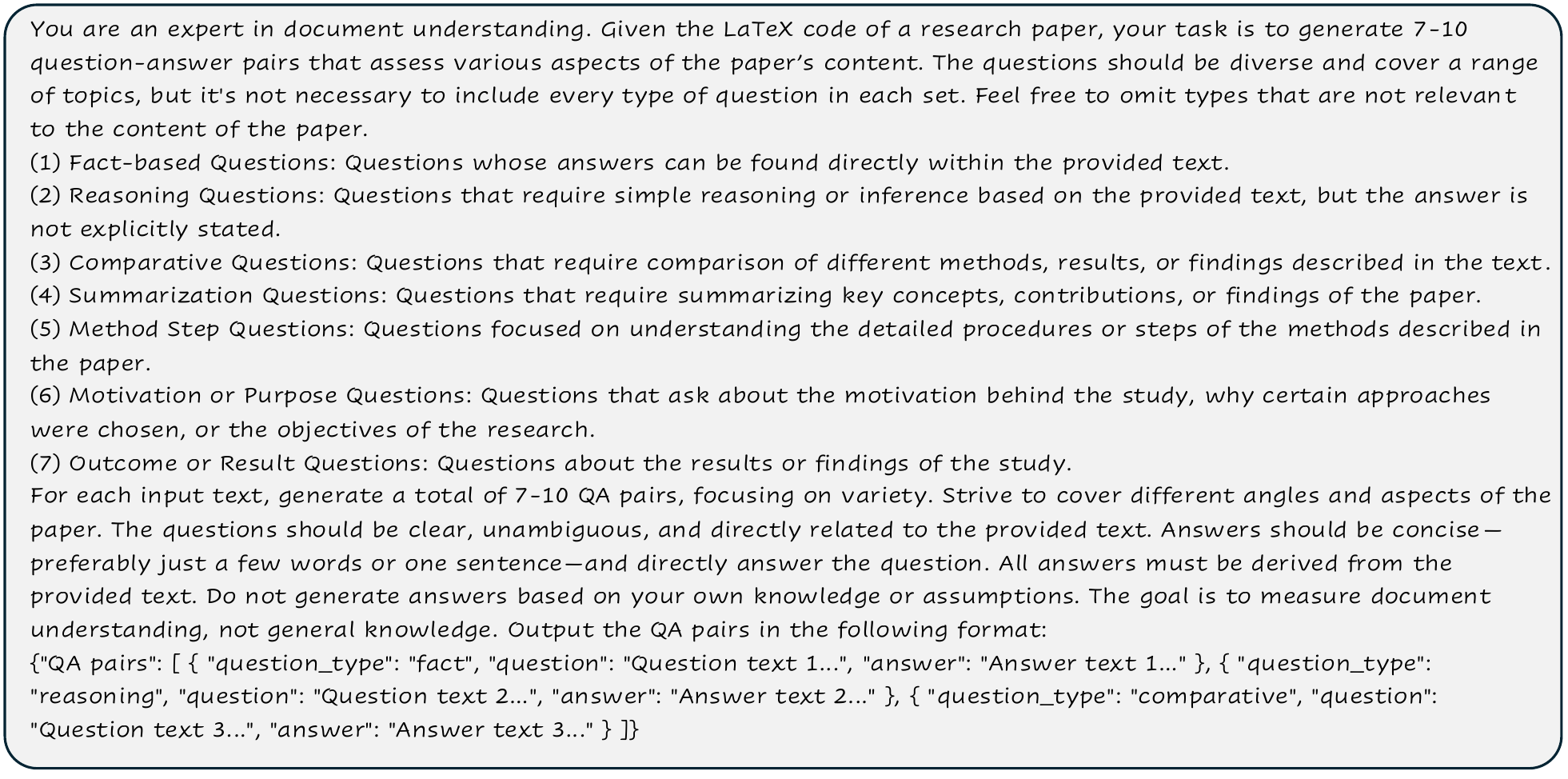}
\caption{Prompt for annotating \AnnotationData.}
\label{fig:generata-arxiv-data}
\end{figure*}

\begin{figure*}[t]
\centering
\includegraphics[width=\textwidth]{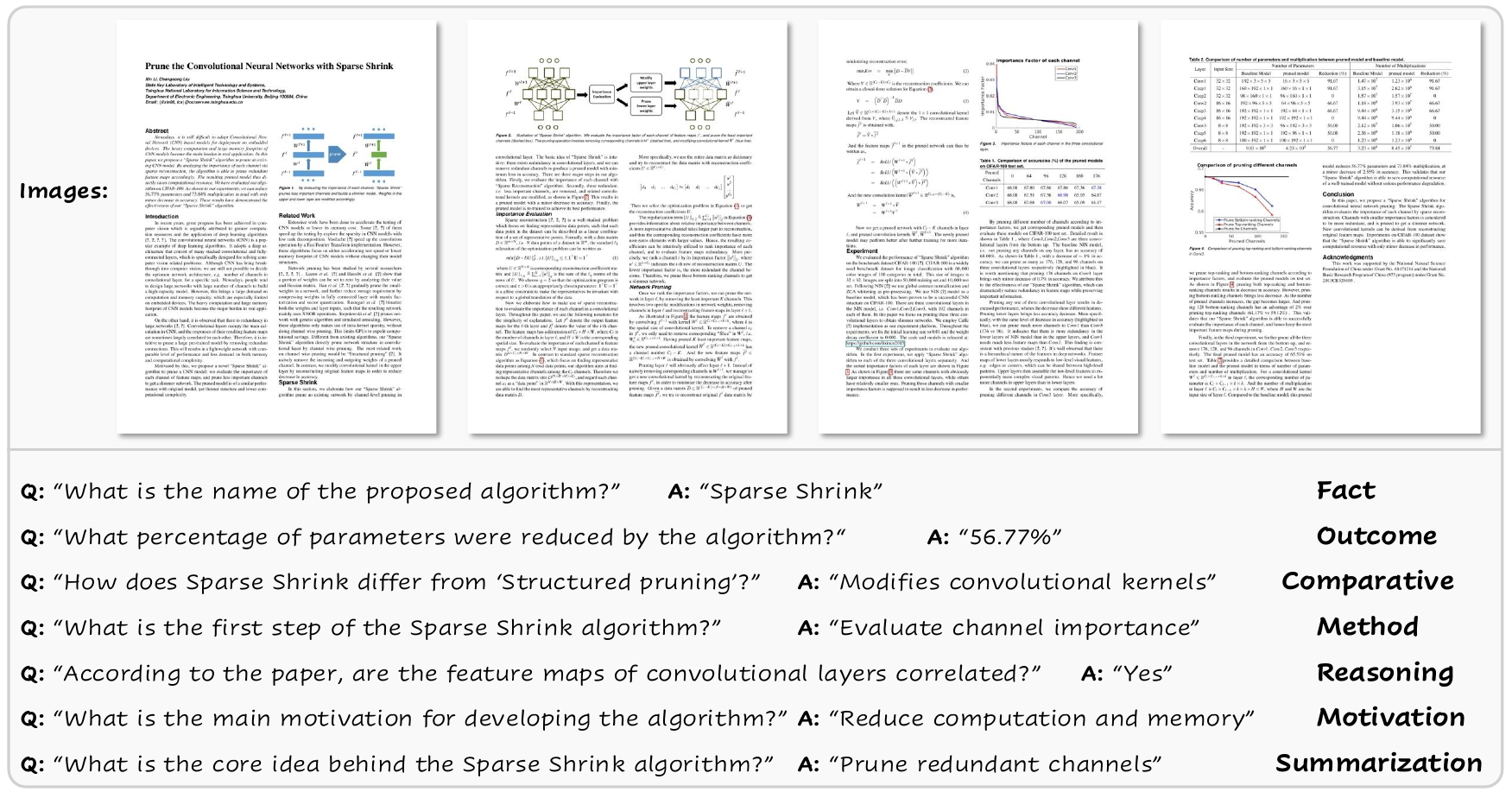}
\caption{Specific annotation examples of the ArxivFullQA benchmark.}
\label{fig:qa-case}
\end{figure*}

The prompt used to annotate the ArxivFullQA evaluation benchmark is shown in Figure~\ref{fig:generata-arxiv-data}.
Examples of the seven types of generated question-answer pairs—including factual, reasoning, comparison, summary, procedural, motivation, and result—are illustrated in Figure~\ref{fig:qa-case}.

\section{Page Selection Formats}
In our experiments, we employ several types of page selection formats (PSFs) for different purposes. Specifically, the PSFs illustrated in Table~\ref{tab:system prompt 1}, Table~\ref{tab:system prompt 2}, and Table~\ref{tab:system prompt no evi} are used in our ablation studies.
PSF 1 (Table~\ref{tab:system prompt 1}) prompts the model to generate evidence pages as a list of image indices, requiring it to internally identify and locate the relevant visual pages.
PSF 2 (Table~\ref{tab:system prompt 2}) adopts a similar True/False format to our main method (PSF 3), but explicitly provides the total number of input images, potentially reducing the difficulty of evidence localization.
In contrast, the PSF shown in Table~\ref{tab:system prompt no evi} omits any evidence-page guidance and serves as the baseline setting in GRPO without support for coarse-to-fine reasoning.

\begin{table}[htbp]
\centering
\begin{tabular}{p{0.95\linewidth}}
\toprule
\textbf{System:} 
You will be given one or more images along with a question. Your task is to understand the visual content and answer the question. First, think carefully about the question and present your reasoning in \texttt{<think>} and \texttt{</think>}. Next, identify all images that contain the necessary evidence to support your answer, and list their page numbers in \texttt{<evidence\_page>} and \texttt{</evidence\_page>}, using integers separated by commas (e.g., 2 or 1, 3, 5). Finally, provide your answer in \texttt{<answer>} and \texttt{</answer>}. The answer should be one or more words or phrases.

\textbf{User:} {\texttt{prompt}}. \textbf{Assistant:} \\
\bottomrule
\end{tabular}
\caption{The first page selection format of EviGRPO (PSF-1). {\texttt{prompt}} will be replaced with the specific question.}
\label{tab:system prompt 1}
\end{table}

\begin{table}[htbp]
\centering
\begin{tabular}{p{0.95\linewidth}}
\toprule
\textbf{System:} 
You will be given \texttt{\{CNT\}} images and a question.  Your task is to understand the visual content and answer the question. First, think carefully about the question and present your reasoning in \texttt{<think>} and \texttt{</think>}. Next, determine whether each page contains relevant evidence to answer the question. Provide your judgment in \texttt{<evidence\_page>} and \texttt{</evidence\_page>} using a comma-separated sequence of T (True) or F (False), one for each page, in order (e.g., T, F, T, F). Finally, provide your answer in \texttt{<answer>} and \texttt{</answer>}. The answer should be one or more words or phrases.

\textbf{User:} {\texttt{prompt}}. \textbf{Assistant:} \\
\bottomrule
\end{tabular}
\caption{The second page selection format of EviGRPO (PSF-2). {\texttt{prompt}} will be replaced with the specific question.}
\label{tab:system prompt 2}
\end{table}

\begin{table}[htbp]
\centering
\begin{tabular}{p{0.95\linewidth}}
\toprule
\textbf{System:} 
You will be given one or more images along with a question. Your task is to understand the visual content and answer the question. First, think carefully about the question and present your reasoning in \texttt{<think>} and \texttt{</think>}. Next, provide your answer in \texttt{<answer>} and \texttt{</answer>}. The answer should be one or more words or phrases.

\textbf{User:} {\texttt{prompt}}. \textbf{Assistant:} \\
\bottomrule
\end{tabular}
\caption{The system prompt of GRPO. {\texttt{prompt}} will be replaced with the specific question.}
\label{tab:system prompt no evi}
\end{table}

\end{document}